\documentclass[runningheads]{llncs}
\usepackage{graphicx}
\usepackage{amsmath,amssymb} 
\usepackage{color}
\usepackage[width=122mm,left=12mm,paperwidth=146mm,height=193mm,top=12mm,paperheight=217mm]{geometry}
\begin{document}
\pagestyle{headings}
\mainmatter

\title{Exploiting temporal information for 3D human pose estimation} 

\titlerunning{Exploiting temporal information for 3D human pose estimation}

\authorrunning{Mir Rayat Imtiaz Hossain, James J. Little}

\author{Mir Rayat Imtiaz Hossain, James J. Little}


\institute{Department of Computer Science,\\
	University of British Columbia\\
	\email{ \{rayat137,little\}@cs.ubc.ca}
}

\maketitle

\begin{abstract}
 In this work, we address the problem of 3D human pose estimation from a sequence of 2D human poses. Although the recent success of deep networks has led many state-of-the-art methods for 3D pose estimation to train deep networks end-to-end to predict from images directly, the top-performing approaches have shown the effectiveness of dividing the task of 3D pose estimation into two steps: using a state-of-the-art 2D pose estimator to estimate the 2D pose from images and then mapping them into 3D space. They also showed that a low-dimensional representation like 2D locations of a set of joints can be discriminative enough to estimate 3D pose with high accuracy. However, estimation of 3D pose for individual frames leads to temporally incoherent estimates due to independent error in each frame causing jitter. Therefore, in this work we utilize the temporal information across a sequence of 2D joint locations to estimate a sequence of 3D poses. We designed a sequence-to-sequence network composed of layer-normalized LSTM units with shortcut connections connecting the input to the output on the decoder side and imposed temporal smoothness constraint during training. We found that the knowledge of temporal consistency improves the best reported result on Human3.6M dataset by approximately $12.2\%$ and helps our network to recover temporally consistent 3D poses over a sequence of images even when the 2D pose detector fails. 

\keywords{3D human pose; sequence-to-sequence Networks; layer normalized LSTM; residual connections}
\end{abstract}

\section{Introduction}

\begin{figure}[t]
\begin{center}
\includegraphics[width=\linewidth,height=0.35\linewidth]{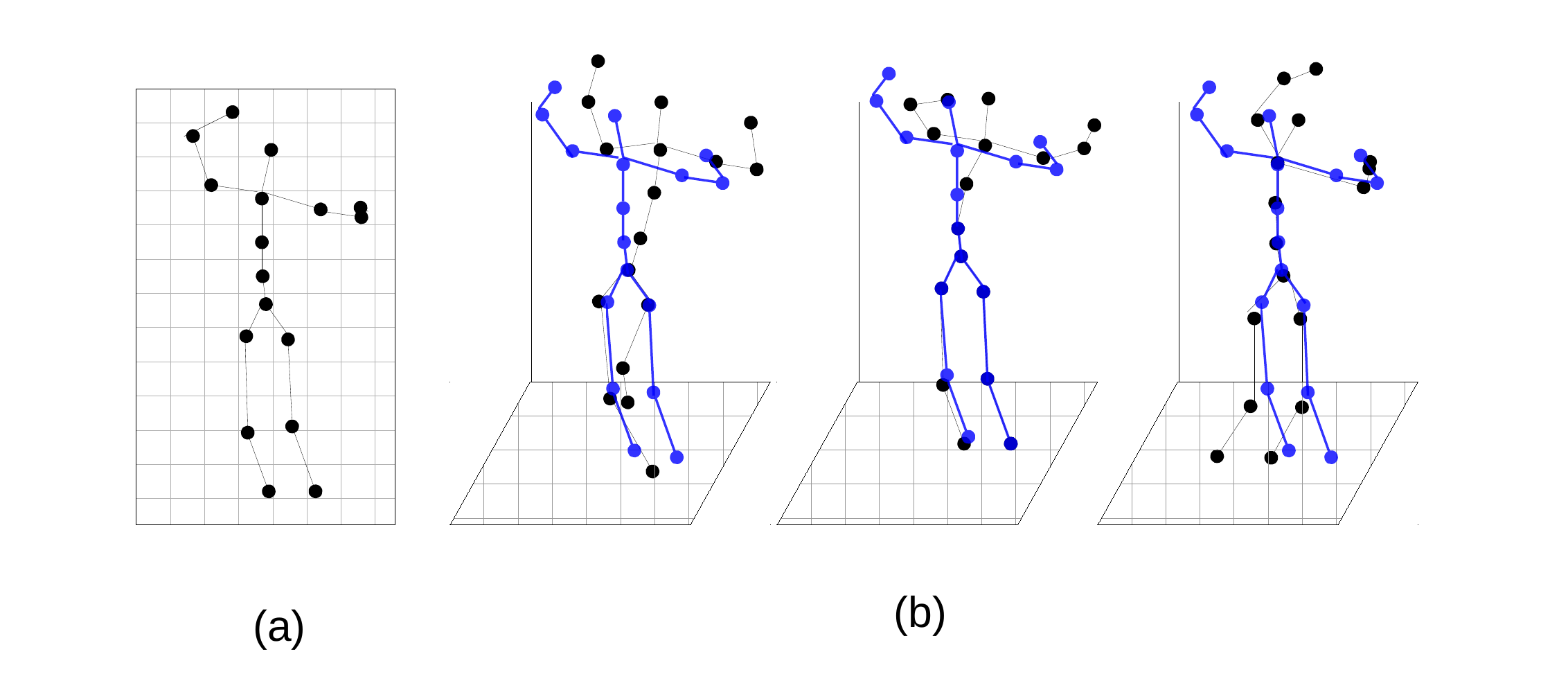}
\end{center}
\caption{\textbf{(a)} 2D position of joints, \textbf{(b)} Different 3D pose interpretations of the same 2D pose. Blue points represent the ground truth 3D locations of joints while the black points indicate other possible 3D interpretations. All these 3D poses project to exactly same 2D pose depending on the position and orientation of the camera projecting them onto 2D plane.}
\vspace{-3mm}
\label{fig:difficulty}   
\end{figure}

The task of estimating 3D human pose from 2D representations like monocular images or videos is an open research problem among the computer vision and graphics community for a long time. An understanding of human posture and limb articulation is important for high level computer vision tasks such as human action or activity recognition, sports analysis, augmented and virtual reality. A 2D representation of human pose, which is considered to be much easier to estimate, can be used for these tasks. However, 2D poses can be ambiguous  because of occlusion and foreshortening. Additionally poses that are totally different can appear to be similar in 2D because of the way they are projected as shown in Figure~\ref{fig:difficulty}. The depth information in 3D representation of human pose makes it free from such ambiguities and hence can improve performance for higher level tasks.  Moreover, 3D pose can be very useful in computer animation, where the articulated pose of a person in 3D can be used to accurately model human posture and movement. 
However, 3D pose estimation is an ill-posed problem because of the inherent ambiguity in back-projecting a 2D view of an object to the 3D space maintaining its structure. Since the 3D pose of a person can be projected in an infinite number of ways on a 2D plane, the mapping from a 2D pose to 3D is not unique. Moreover, obtaining a dataset for 3D pose is difficult and expensive. Unlike the 2D pose datasets where the users can manually label the keypoints by mouse clicks, 3D pose datasets require a complicated laboratory setup with motion capture sensors and cameras. Hence, there is a lack of motion capture datasets for images in-the-wild.

\begin{figure*}
\centering
\includegraphics[width=\linewidth,height=0.45\linewidth]{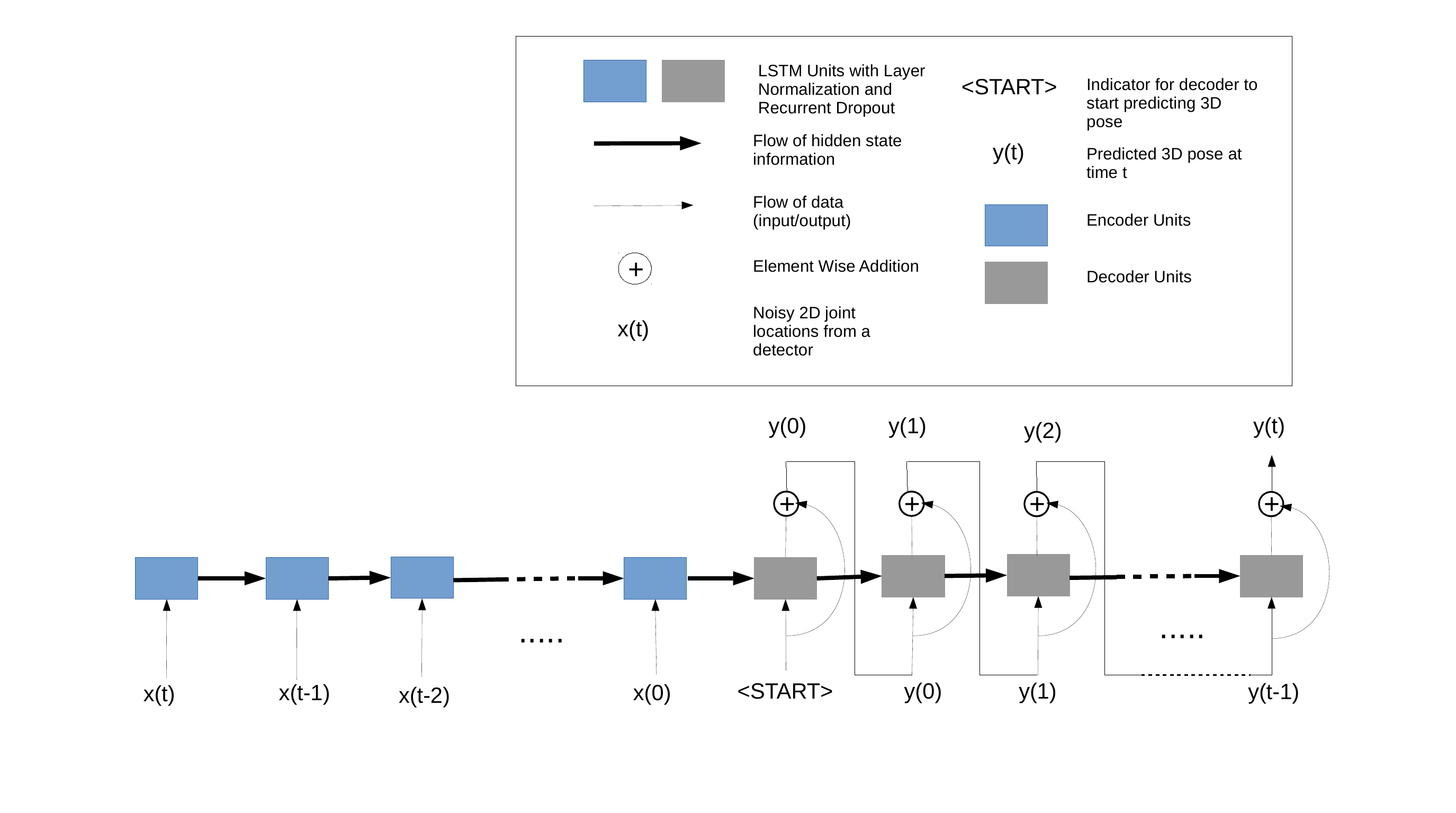}
\caption{Our model. It is a sequence-to-sequence network~\cite{sutskever2014sequence} with residual connections on the decoder side. The encoder encodes the information of a sequence of 2D poses of length t in its final hidden state. The final hidden state of the encoder is used to initialize the hidden state of decoder. The $\langle START \rangle$ symbol tells the decoder to start predicting 3D pose from the last hidden state of the encoder.  Note that the input sequence is reversed as suggested by Sutskever et al.~\cite{sutskever2014sequence}.  The decoder essentially learns to predict the 3D pose at time $(t)$ given the 3D pose at time $(t-1)$. The residual connections help the decoder to learn the perturbation from the previous time step.}
\vspace{-3mm}
\label{fig:network_model}   
\end{figure*}

Over the years, different techniques have been used to address the problem of 3D pose estimation. Earlier methods used to focus on extracting features, invariant to factors such as background scenes, lighting, and skin color from images and mapping them into 3D human pose~\cite{AgarwalT04,bb93102,bb41073,bb93266}. With the success of deep networks, recent methods tend to focus on training a deep convolutional neural network (CNN) end-to-end to estimate 3D poses from images directly~\cite{tekin2016structured,volumetric,li20143d,mehta2016monocular,zhou2016deep,mehta2017vnect,niemonocular,linCVPR17RPSM,park20163d,sun2017compositional,tekin2017learning}. Some approaches divided the 3D pose estimation task into first predicting the joint locations in 2D using 2D pose estimators~\cite{cpm,stacked-hourglass} and then back-projecting them to estimate the 3D joint locations~\cite{ramakrishna2012reconstructing,zhou2016sparseness,akhter2015pose,bogo2016keep,distance-matrix,JMartinez:ICCV:2017}. These results suggest the effectiveness of decoupling the task of 3D pose estimation where 2D pose estimator abstracts the complexities in the image. In this paper, we also adopt the decoupled approach to 3D pose estimation. However, predicting 3D pose for each frame individually can lead to jitter in videos because the errors in each frame are independent of each other. Therefore, we designed a sequence-to-sequence network~\cite{sutskever2014sequence} with shortcut connections on the decoder side~\cite{he2016deep} that predicts a sequence of temporally consistent 3D poses given a sequence of 2D poses. Each unit of our network is a Long Short-Term Memory (LSTM)~\cite{hochreiter1997long} unit with layer normalization~\cite{ba2016layer} and recurrent dropout~\cite{zaremba2014recurrent}. We also imposed a temporal smoothness constraint on the predicted 3D poses during training to ensure that our predictions are smooth over a sequence. 

Our network achieves the state-of-the-art result on the Human3.6M dataset improving the previous best result by approximately $12.2\%$. We also obtained the lowest error for every action class in Human3.6M dataset~\cite{h36m_pami}. Moreover, we observed that our network predicted meaningful 3D poses on Youtube videos, even when the detections from the 2D pose detector were extremely noisy or meaningless. This shows the effectiveness of using temporal information. In short our contributions in this work are:
\begin{itemize}
\item Designing an efficient sequence-to-sequence network that achieves the state-of-the-art results for every action class of Human3.6M dataset~\cite{h36m_pami} and can be trained very fast.
\item Exploiting the ability of sequence-to-sequence networks to take into account the  events in the past, to predict temporally consistent 3D poses. 
\item Effectively imposing temporal consistency constraint on the predicted 3D poses during training so that the errors in the predictions are distributed smoothly over the sequence. 
\item Using only the previous frames to understand temporal context so that it can be deployed online and real-time.
\end{itemize}


\section{Related Work}
\paragraph{Representation of 3D pose}
Both model-based and model-free representations of 3D human pose have been used in the past. The most common model-based representation is a skeleton defined by a kinematic tree of a set of joints, parameterized by the offset and rotational parameters of each joint relative to its parent. Several 3D pose methods have used this representation~\cite{barron_2001,ParameswaranC04,bogo2016keep,zhou2016deep}. Others model 3D pose as a sparse linear combination of an over-complete dictionary of basis poses~\cite{akhter2015pose,zhou2016sparseness,ramakrishna2012reconstructing}. However, we have chosen a model-free representation of 3D pose, where a 3D pose is simply a set of 3D joint locations relative to the root node like several recent approaches~\cite{JMartinez:ICCV:2017,distance-matrix,li20143d,mehta2016monocular}. This representation is much simpler and low-dimensional.

\paragraph{Estimating 3D pose from 2D joints}

Lee and Chen~\cite{Chen85b} were the first to infer 3D joint locations from their 2D projections given the bone lengths using a  binary decision tree where each branch corresponds to two possible states of a joint relative to its parent. Jiang~\cite{jiang20103d} used the 2D joint locations to estimate a set of hypothesis 3D poses using Taylor's algorithm~\cite{taylor2000reconstruction} and used them to query a large database of motion capture data to find the nearest neighbor. Gupta et al.~\cite{gupta20143dpose} and Chen and Ramanan~\cite{chen20163d} also used this idea of using the detected 2D pose to query a large database of exemplar poses to find the nearest nearest neighbor 3D pose. Another common approach to estimating 3D joint locations given the 2D pose is to separate the camera pose variability from the intrinsic deformation of the human body, the latter of which is modeled by learning an over-complete dictionary of basis 3D poses from a large database of motion capture data~\cite{ramakrishna2012reconstructing,zhou2016sparseness,bogo2016keep,akhter2015pose,wang2014robust}. A valid 3D pose is defined by a sparse linear combination of the bases and by transforming the points using transformation matrix representing camera extrinsic parameters. Moreno-Nouguer~\cite{distance-matrix} used the pair-wise distance matrix of 2D joints to learn a distance matrix for 3D joints, which they found invariant up to a rigid similarity transform with the ground truth 3D and used multi-dimensional scaling (MDS) with pose-priors to rule out the ambiguities. Martinez et al.~\cite{JMartinez:ICCV:2017} designed a fully connected network with shortcut connections every two linear layers to estimate 3D joint locations relative to the root node in the camera coordinate space.

\paragraph{Deep network based methods}
With the success of deep networks, many have designed networks that can be trained end-to-end to predict 3D poses from images directly~\cite{volumetric,li20143d,tekin2016structured,park20163d,mehta2016monocular,sun2017compositional,zhou2016deep,varol_2017,RogezS16,tome2017lifting}. Li et al.~\cite{li20143d} and Park et al.~\cite{park20163d} designed CNNs to jointly predict 2D and 3D poses. Mehta et al.~\cite{mehta2016monocular} and Sun et al.~\cite{sun2017compositional} used transfer learning to transfer the knowledge learned for 2D human pose estimation to the task of 3D pose estimation. Pavlakos et al.~\cite{volumetric} extended the stacked-hourglass network~\cite{stacked-hourglass} originally designed to predict 2D heatmaps of each joint to make it predict 3D volumetric heatmaps. Tome  et al.~\cite{tome2017lifting} also extended a 2D pose estimator called Convolutional Pose Machine (CPM)~\cite{cpm} to make it predict 3D pose. Rogesz and Schmid~\cite{RogezS16} and Varol et al.~\cite{varol_2017} augmented the training data with synthetic images and trained CNNs to predict 3D poses from real images. Sun et al.~\cite{sun2017compositional} designed a unified network that can regress both 2D and 3D poses at the same time given an image. Hence during training time, in-the-wild images which do not have any ground truth 3D poses can be combined with the data with ground truth 3D poses. A similar idea of exploiting in-the-wild images to learn pose structure was used by Fang et al.~\cite{fang2017learning}. They learned a pose grammar that encodes the possible human pose configurations.

\paragraph{Using temporal information}
Since estimating poses for each frame individually leads to incoherent and jittery predictions over a sequence, many approaches tried to exploit temporal information~\cite{andriluka2010monocular,tekin2016direct,zhou2016sparseness,du2016marker,mehta2017vnect}. Andriluka et al.~\cite{andriluka2010monocular} used tracking-by-detection to associate 2D poses detected in each frame individually and used them to retrieve 3D pose. Tekin et al.~\cite{tekin2016direct} used a CNN to first align bounding boxes of successive frames so that the person in the image is always at the center of the box and then extracted 3D HOG features densely over the spatio-temporal volume from which they regress the 3D pose of the central frame. Mehta et al.~\cite{mehta2017vnect} implemented a real-time system for 3D pose estimation that applies temporal filtering across 2D and 3D poses from previous frames to predict a temporally consistent 3D pose. Lin et al.~\cite{linCVPR17RPSM} performed a multi-stage sequential refinement using LSTMs to predict 3D pose sequences using previously predicted 2D pose representations and 3D pose. We focus on predicting temporally consistent 3D poses by learning the temporal context of a sequence using a form of sequence-to-sequence network. Unlike Lin et al.~\cite{linCVPR17RPSM} our method does not need multiple stages of refinement. It is simpler and requires fewer parameters to train, leading to much improved performance. 

\section{Our Approach}
\paragraph{Network Design}
We designed a sequence-to-sequence network with LSTM units and residual connections on the decoder side to predict a temporally coherent sequence of 3D poses given a sequence of 2D joint locations. Figure~\ref{fig:network_model} shows the architecture of our network. The motivation behind using a sequence-to-sequence network comes from its application on the task of Neural Machine Translation (NMT) by Sutskever  et al.~\cite{sutskever2014sequence}, where their model translates a sentence in one language to a sentence in another language e.g. English to French. In a language translation model, the input and output sentences can have different lengths. Although our case is analogous to the NMT, the input and output sequences always have the same length while the input vectors to the encoder and decoder have different dimensions.

The encoder side of our network takes a sequence of 2D poses and encodes them in a fixed size high dimensional vector in the hidden state of its final LSTM unit. Since the LSTMs are excellent in memorizing events and information from the past, the encoded vector stores the 2D pose information of all the frames. The initial state of the decoder is initialized by the final state of the encoder. A $\langle START \rangle$ token is passed as initial input to the decoder, which in our case is a vector of ones, telling it to start decoding. Given a 3D pose estimate $y_t$ at a time step $t$ each decoder unit predicts the 3D pose for next time step $y_{t+1}$. Note that the order of the input sequence is reversed as recommended by Sutskever et al.~\cite{sutskever2014sequence}. The shortcut connections on the decoder side cause each decoder unit to estimate the amount of perturbation in the 3D pose from the previous frame instead of having to estimate the actual 3D pose for each frame. As suggested by He et al.~\cite{he2016deep}, such a mapping is easier to learn for the network.

We use layer normalization~\cite{ba2016layer} and recurrent dropout~\cite{zaremba2014recurrent} to regularize our network. Ba et al.~\cite{ba2016layer} came up with the idea of layer normalization which estimates the normalization statistics (mean and standard deviation) from the summed inputs to the recurrent neurons of hidden layer on a \emph{single} training example to regularize the RNN units. Similarly, Zaremba et al.~\cite{zaremba2014recurrent} proposed the idea of applying dropout only on the non-recurrent connections of the network with a certain probability $p$ while always keeping the recurrent connections intact because they are necessary for the recurrent units to remember the information from the past.  
\vspace{-5mm}
\paragraph{Loss function}
Given a sequence of 2D joint locations as input, our network predicts a sequence of 3D joint locations relative to the root node (central hip). We predict each 3D pose in the camera coordinate space instead of predicting them in an arbitrary global frame as suggested by Martinez et al.~\cite{JMartinez:ICCV:2017}.

We impose a temporal smoothness constraint on the predicted 3D joint locations to ensure that the prediction of each joint in one frame does not differ too much from its previous frame. Because the 2D pose detectors work on individual frames, even with the minimal movement of the subject in the image, the detections from successive frames may vary, particularly for the joints which move fast or are prone to occlusion. Hence, we made an assumption that the subject does not move too much in successive frames given the frame rate is high enough. Therefore, we added the L2 norm of the first order derivative on the 3D joint locations with respect to time to our loss function during training. This constraint helps us to estimate 3D poses reliably even when the 2D pose detector fails for a few frames within the temporal window without any post-processing.

Empirically we found that certain joints are  more difficult to estimate accurately e.g. wrist, ankle, elbow compared to others. To address this issue, we partitioned the joints into three disjoint sets $\mathbf{torso\_head}$, $\mathbf{limb\_leg}$ and $\mathbf{limb\_arm}$ based on their contribution to overall error. We observed that the joints connected to the torso and the head e.g. hips, shoulders, neck are always predicted with high accuracy compared to those joints belonging to the limbs and therefore put them in the set $\mathbf{torso\_head}$. The joints of the limbs, especially the joints on the arms, are always more difficult to predict due to their high range of motion and occlusion. We put the knees and the ankles in the set $\mathbf{limb\_leg}$  and the elbow and wrist in $\mathbf{limb\_arm}$. We multiply the derivatives of each set of joints with different scalar values based on their contribution to the overall error.

Therefore our loss function consists of the sum of two separate terms: Mean Squared Error (MSE) of $N$ different sequences of 3D joint locations; and the mean of the L2 norm of the first order derivative of $N$ sequences of 3D joint locations with respect to time, where the joints are divided into three disjoint sets.

The MSE over $N$ sequences, each of $T$ time-steps, of 3D joint locations is given by 

\begin{equation}
\mathcal{\mathbf{L}}(\mathbf{\hat{Y}},\mathbf{Y}) = \frac{1}{NT}\sum_{i=1}^{N}\sum_{t=1}^{T}\left\Vert \mathbf{\hat{Y}_{i,t}} -  \mathbf{{Y}_{i,t}}  \right\Vert^{2}_2. 
\end{equation} Here, $\mathbf{\hat{Y}}$ denotes the estimated 3D joint locations while $\mathbf{Y}$ denotes 3D ground truth. 

The mean of L2 norm of the first order derivative of $N$ sequences of 3D joint locations, each of length $T$, with respect to time is given by

\begin{multline}
\left\Vert\mathbf{\nabla_{t}}\mathbf{\hat{Y}} \right\Vert^{2}_2 = \frac{1}{N(T-1)}  \sum_{i=1}^{N}\sum_{t=2}^{T}\left\{\mathbf{\eta}\left\Vert \mathbf{\hat{Y}_{i,t}^{TH}} -  \mathbf{\hat{Y}_{i,t-1}^{TH}}  \right\Vert^{2}_2 \right. \\ + \mathbf{\rho} \left\Vert \mathbf{\hat{Y}_{i,t}^{LL}} -  \mathbf{\hat{Y}_{i,t-1}^{LL}}  \right\Vert^{2}_2 + \left. \mathbf{\tau}  \left\Vert \mathbf{\hat{Y}_{i,t}^{LA}} -  \mathbf{\hat{Y}_{i,t-1}^{LA}}  \right\Vert^{2}_2 \right\}.
\end{multline} In the above equation, $\mathbf{\hat{Y}^{TH}}$, $\mathbf{\hat{Y}^{LL}}$ and $\mathbf{\hat{Y}^{LA}}$ denotes the predicted 3D locations of joints belonging to the sets $\mathbf{torso\_head}$, $\mathbf{limb\_leg}$ and $\mathbf{limb\_arm}$ respectively. The $\mathbf{\eta}, \mathbf{\rho}$ and $\mathbf{\tau}$ are scalar hyper-parameters to control the significance of the derivatives of 3D locations of each of the three set of joints. A higher weight is assigned to the set of joints which are generally predicted with higher error.  

The overall loss function for our network is given as

\begin{equation}
\mathcal{\mathbf{L}} =\min_{\mathbf{\hat{Y}}}\mathbf{\alpha} \mathbf{L}(\mathbf{\hat{Y}},\mathbf{Y})  +\mathbf{\beta }\left\Vert\mathbf{\nabla_{t}}\mathbf{\hat{Y}} \right\Vert^{2}_2. 
\end{equation}Here $\alpha$ and $\beta$ are scalar hyper-parameters regulating the importance of each of the two terms in the loss function. 

\section{Experimental Evaluation}

\begin{table}
\footnotesize
\centering
\tabcolsep=0.6mm
\scalebox{0.58}{
\begin{tabular}{@{}lrrrrrrrrrrrrrrrr@{}}
\hline
Protocol \#1 & Direct. & Discuss & Eating & Greet & Phone & Photo & Pose & Purch. & Sitting & SitingD & Smoke & Wait & WalkD & Walk & WalkT & Avg\\
\hline
LinKDE ~\cite{h36m_pami} (SA)  & 132.7 & 183.6 & 132.3 & 164.4 & 162.1 & 205.9 & 150.6 & 171.3 & 151.6 & 243.0 & 162.1 & 170.7 & 177.1 & 96.6 & 127.9 & 162.1\\
Tekin et al~\cite{tekin2016direct} (SA) & 102.4 & 147.2 & 88.8 & 125.3 & 118.0 & 182.7 & 112.4 & 129.2 & 138.9 & 224.9 & 118.4 & 138.8 & 126.3 & 55.1 & 65.8 & 125.0\\
Zhou et al~\cite{zhou2016sparseness} (MA) & 87.4 & 109.3 & 87.1 & 103.2 & 116.2 & 143.3 & 106.9 & 99.8 & 124.5 & 199.2 & 107.4 & 118.1 & 114.2 & 79.4 & 97.7 & 113.0\\
Park et al~\cite{park20163d} (SA) & 100.3 & 116.2 & 90.0 & 116.5 & 115.3 & 149.5 & 117.6 & 106.9 &  137.2 & 190.8 & 105.8 & 125.1 & 131.9 & 62.6 & 96.2 & 117.3\\
Nie et al~\cite{niemonocular} (MA)   & 90.1 & 88.2 & 85.7 & 95.6 & 103.9 & 103.0 & 92.4 & 90.4 & 117.9 & 136.4 & 98.5 & 94.4 & 90.6 & 86.0 & 89.5 & 97.5\\
Mehta et al~\cite{mehta2016monocular} (MA)   & 57.5 & 68.6 & 59.6 & 67.3 & 78.1 & 82.4 & 56.9  & 69.1 & 100.0 & 117.5 & 69.4 & 68.0 & 76.5 & 55.2 & 61.4 & 72.9\\
Mehta et al~\cite{mehta2017vnect} (MA)   & 62.6 & 78.1 & 63.4 & 72.5 & 88.3 & 93.8 & 63.1  & 74.8 & 106.6 & 138.7 & 78.8 & 73.9 & 82.0 & 55.8 & 59.6 & 80.5\\
Lin et al~\cite{linCVPR17RPSM} (MA) & 58.0 & 68.2 & 63.3 & 65.8 & 75.3 & 93.1 & 61.2  & 65.7 & 98.7 & 127.7 & 70.4 & 68.2 & 72.9 & 50.6 & 57.7 & 73.1\\
Tome et al~\cite{tome2017lifting} (MA)  & 65.0 & 73.5 & 76.8 & 86.4 & 86.3 & 110.7 & 68.9  & 74.8 & 110.2 & 173.9 & 84.9 & 85.8 & 86.3 & 71.4 & 73.1 & 88.4\\
Tekin et al~\cite{tekin2017learning} & 54.2 & 61.4 & 60.2 & 61.2 & 79.4 & 78.3 & 63.1 & 81.6 & 70.1 & 107.3 & 69.3 & 70.3 & 74.3 & 51.8 & 63.2 & 69.7\\
Pavlakos et al~\cite{volumetric} (MA) & 67.4 & 71.9 & 66.7 & 69.1 & 72.0 & 77.0 & 65.0 & 68.3 & 83.7 & 96.5 & 71.7 & 65.8 & 74.9 & 59.1 & 63.2 & 71.9\\
Martinez et al.~\cite{JMartinez:ICCV:2017} (MA) & 51.8&  56.2&	58.1&	59.0&	69.5&	78.4&	55.2&	58.1&	74.0&	94.6&	62.3&	59.1&	65.1&	49.5&	52.4&	62.9\\ 
Fang et al.~\cite{fang2017learning} (MA) 17j  & \underline{50.1} & \underline{54.3} & 57.0 & 57.1 & 66.6 & 73.3 & 53.4 & 55.7 & 72.8 & 88.6 & \underline{60.3} & 57.7 & 62.7 & 47.5 & \underline{50.6} & 60.4\\
Sun et al.~\cite{sun2017compositional} (MA) 17j  & 52.8 & 54.8 & \underline{54.2} &\underline {54.3} & \underline{61.8} & \underline{67.2} & \underline{53.1} & \underline{53.6} & \underline{71.7} & \underline{86.7} & 61.5 & \underline{53.4} & \underline{61.6}  & \underline{47.1} & 53.4 & \underline{59.1}\\
\hline
Baseline 1 (~\cite{JMartinez:ICCV:2017} + median filter)  & 51.8& 	55.3& 	59.1& 	58.5& 	66.4& 	79.2& 	54.7& 	55.8& 	73.2& 	89.0& 	61.6& 	59.5& 	65.9& 	49.5& 	53.5& 	62.2\\
Baseline 2 (~\cite{JMartinez:ICCV:2017} + mean filter)  & 50.9 &  54.9& 58.2 &	57.9& 65.6 &	78.9&	53.7&	55.8&	73.5&	89.9& 60.9 & 59.2 &	65.1& 49.2 &	52.8& 61.8 \\ 
\textbf{Our network (MA)} & \bf{44.2}&  \bf{46.7}&	\bf{52.3}&	\bf{49.3}&	\bf{59.9}&	\bf{59.4}&	\bf{47.5}&	\bf{46.2}&	\bf{59.9}&	\bf{65.6}&	\bf{55.8}&	\bf{50.4}&	\bf{52.3}&	\bf{43.5}&	\bf{45.1}&	\bf{51.9}\\

\hline
Martinez et al.~\cite{JMartinez:ICCV:2017} (GT) (MA) & 37.7& 	44.4& 	40.3& 	42.1& 	48.2& 	54.9& 	44.4& 	42.1& 	54.6& 	58.0& 	45.1& 	46.4& 	47.6& 	36.4& 	40.4& 	45.5\\
\textbf{Our network (GT) (MA)} & \it{35.2}&  \it{40.8}&	\it{37.2}&	\it{37.4}&	\it{43.2}&	\it{44.0}&	\it{38.9}&	\it{35.6}&	\it{42.3}&	\it{44.6}&	\it{39.7}&	\it{39.7}&	\it{40.2}&	\it{32.8}&	\it{35.5}&	\it{39.2}\\
\hline
\end{tabular}}
\vspace{3mm}
\caption{Results showing the errors action-wise on Human3.6M~\cite{h36m_pami} under Protocol \#1 (no rigid alignment or similarity transform applied in post-processing). Note that our results reported here are for sequence of length 5. SA indicates that a model was trained for each action, and MA indicates that a single model was trained for all actions. GT indicates that the network was trained on ground truth 2D pose. The bold-faced numbers represent the best result while underlined numbers represent the second best.}
\label{tab:protocol_1}
\vspace{-5mm}
\end{table}

\paragraph{Datasets and protocols} 

We perform quantitative evaluation on the Human 3.6M~\cite{h36m_pami} dataset and on the HumanEva dataset~\cite{heva}. Human 3.6M, to the best of our knowledge, is the largest publicly available dataset for human 3D pose estimation. The dataset contains 3.6 million images of 7 different professional actors performing 15 everyday activities like walking, eating, sitting, making a phone call. The dataset consists of 2D and 3D joint locations for each corresponding image. Each video is captured using 4 different calibrated high resolution cameras. In addition to 2D and 3D pose ground truth, the dataset also provides ground truth for bounding boxes, the camera parameters, the body proportion of all the actors and high resolution body scans or meshes of each actor. HumanEva, on the other hand, is a much smaller dataset. It has been largely used to benchmark previous work over the last decade. Most of the methods report results on two different actions and on three actors. For qualitative evaluation, we used the some videos from Youtube and the Human3.6M dataset.

We follow the standard protocols of the Human3.6M dataset used in the literature. We used subjects 1, 5, 6, 7, and 8 for training, and subjects 9 and 11 for testing and the error is evaluated on the predicted 3D pose without any transformation. We refer this as protocol \#1. Another common approach used by many to evaluate their methods is to align the predicted 3D pose with the ground truth using a similarity transformation (Procrustes analysis). We refer this as protocol \#2. We use the average error per joint in millimeters between the estimated and the ground truth 3D pose relative to the root node as the error metric. For the HumanEva dataset, we report results on each subject and action separately after performing rigid alignment with the ground truth data, following the protocol used by the previous methods. 

\begin{table*}
\footnotesize
\tabcolsep=0.6mm
\vspace{-3mm}
\centering
\scalebox{0.58}{
\begin{tabular}{@{}lrrrrrrrrrrrrrrrr@{}}
\hline
Protocol \#2 & Direct. & Discuss & Eating & Greet & Phone & Photo & Pose & Purch. & Sitting & SitingD & Smoke & Wait & WalkD & Walk & WalkT & Avg\\
\hline
Akhter \& Black~\cite{akhter2015pose}* (MA) 14j & 199.2 & 177.6 & 161.8 & 197.8 & 176.2 & 186.5 & 195.4 & 167.3 & 160.7 & 173.7 & 177.8 & 181.9 & 176.2 & 198.6 & 192.7 & 181.1\\
Ramakrishna et al~\cite{ramakrishna2012reconstructing}* (MA) 14j & 137.4 & 149.3 & 141.6 & 154.3 & 157.7 & 158.9 & 141.8 & 158.1 & 168.6 & 175.6 & 160.4 & 161.7 & 150.0 & 174.8 & 150.2 & 157.3\\
Zhou et al~\cite{zhou2016sparseness}* (MA) 14j & 99.7 & 95.8 & 87.9 & 116.8 & 108.3 & 107.3 & 93.5 & 95.3 & 109.1 & 137.5 & 106.0 & 102.2 & 106.5 & 110.4 & 115.2 & 106.7\\
Rogez et al~\cite{mehta2016monocular} (MA)   & -- & -- & -- &-- & -- & -- & --  & -- & -- & -- & -- & -- & -- & -- &-- & 87.3\\
Nie et al~\cite{niemonocular} (MA)   & 62.8 & 69.2 & 79.6 & 78.8 & 80.8 & 86.9 & 72.5 & 73.9 & 96.1 & 106.9 & 88.0 & 70.7 & 76.5 & 71.9 & 76.5 & 79.5\\
Mehta et al~\cite{mehta2016monocular} (MA) 14j  & -- & -- & -- & -- & -- & -- & -- & -- & -- & -- & -- & -- & -- & -- & -- & 54.6\\
Bogo et al~\cite{bogo2016keep} (MA) 14j & 62.0 & 60.2 & 67.8 & 76.5 & 92.1 & 77.0 & 73.0 & 75.3 & 100.3 & 137.3 & 83.4 & 77.3 & 86.8 & 79.7 & 87.7 & 82.3\\
Moreno-Noguer~\cite{distance-matrix} (MA) 14j & 66.1 & 61.7 & 84.5 & 73.7 & 65.2 & 67.2 & 60.9 & 67.3 & 103.5 & 74.6 & 92.6 & 69.6 & 71.5 & 78.0 & 73.2 & 74.0\\
Tekin et al~\cite{tekin2017learning} (MA) 17j  & -- & -- & -- & -- & -- & -- & -- & -- & -- & -- & -- & -- & -- & -- & -- & 50.1\\
Pavlakos et al~\cite{volumetric} (MA) 17j & -- & -- & -- & -- & -- & -- & -- & -- & -- & -- & -- & -- & -- & -- & -- & 51.9\\
Martinez et al.~\cite{JMartinez:ICCV:2017} (MA) 17j & 39.5 & 43.2 &	46.4&	47.0&	51.0&56.0&	41.4&	40.6&	56.5&	69.4&	49.2&	45.0&	49.5&	38.0&	43.1&	47.7\\
Fang et al.~\cite{fang2017learning} (MA) 17j  & \underline{38.2} & \underline{41.7} & \underline{43.7} & \underline{44.9}  & \underline{48.5}  & \underline{55.3} & \underline{40.2} & \underline{38.2} & \underline{54.5} & \underline{64.4} & \underline{47.2} & \underline{44.3} & \underline{47.3} & \underline{36.7} & \underline{41.7} & \underline{45.7}\\
\hline
Baseline 1 (~\cite{JMartinez:ICCV:2017} + median filter)  & 44.1& 	46.3& 	49.6& 	50.3& 	53.2& 	60.9& 	43.7& 	43.5& 	61.2& 	74.4& 	53.0& 	48.6& 	54.7& 	43.0& 	48.5& 	51.7\\
Baseline 2 (~\cite{JMartinez:ICCV:2017} + mean filter)  & 43.1 &  45.0& 48.8 &	49.0& 52.1 &	59.4&	43.5&	42.4&	59.7&	70.9& 51.2 & 46.9 &	52.4& 40.3 &	46.0& 50.0 \\ 
\textbf{Our network (MA) 17j} & \bf{36.9}&  \bf{37.9}&	\bf{42.8}&	\bf{40.3}&	\bf{46.8}&	\bf{46.7}&	\bf{37.7}&	\bf{36.5}&	\bf{48.9}&	\bf{52.6}&	\bf{45.6}&	\bf{39.6}&	\bf{43.5}&	\bf{35.2}&	\bf{38.5}&	\bf{42.0}\\
\hline
\end{tabular}
}
\vspace{3mm}
\caption{Results showing the errors action-wise on Human3.6M~\cite{h36m_pami} dataset under protocol \#2 (Procrustes alignment to the ground truth in post-processing). Note that the results reported here are for sequence of length 5. The 14j  annotation indicates that the body model considers 14  body joints while 17j means considers 17 body joints. (SA) annotation indicates per-action model while (MA) indicates single model used for all actions. The bold-faced numbers represent the best result while underlined numbers represent the second best. The results of the methods are obtained from the original
papers, except for (*), which were obtained from~\cite{bogo2016keep}.}
\vspace{-5mm}
\label{tab:protocol_2}
\end{table*}

\paragraph{2D detections} 
We fine-tuned a model of stacked-hourglass network~\cite{stacked-hourglass}, initially trained on the MPII dataset~\cite{mpii} (a benchmark dataset for 2D pose estimation), on the images of the Human3.6M dataset to obtain 2D pose estimations for each image. We used the bounding box information provided with the dataset to first compute the center of the person in the image and then cropped a $440 \times 440$ region across the person and resized it to $256 \times 256$. We fine-tuned the network for 250 iterations and used a batch size of 3 and a learning rate of $2.5e-4$.
\paragraph{Baselines} 
Since many of the previous methods are based on single frame predictions, we used two baselines for comparison. To show that our method is much better than naive post processing, we applied a mean filter and a median filter on the 3D pose predictions of Martinez et al.~\cite{JMartinez:ICCV:2017}. We used a window size of 5 frames and a stride length of 1 to apply the filters. Although non-rigid structure from motion (NRSFM) is one of the most general approaches for any 3D reconstruction problem from a sequence of 2D correspondences, we did not use it as a baseline because Zhou et al.~\cite{zhou2016sparseness} did not find NRSFM techniques to be effective for 3D human pose estimation.They found that the NRSFM techniques do not work well with slow camera motion. Since the videos in the Human3.6M dataset~\cite{h36m_pami} are captured by stationary cameras,the subjects in the dataset do not rotate that much to provide alternative views for NRSFM algorithm to perform well. Another reason is that human pose reconstruction is a specialized problem in which constraints from human body structure apply.

\paragraph{Data pre-processing} 
We normalized the 3D ground truth poses, the noisy 2D pose estimates from stacked-hourglass network and the 2D ground truth~\cite{stacked-hourglass} by subtracting the mean and dividing by standard deviation. We do not predict the 3D location of the root joint i.e. central hip joint and hence zero center the 3D joint locations relative to the global position of the root node. To obtain the ground truth 3D poses in camera coordinate space, an inverse rigid body transformation is applied on the the ground truth 3D poses in global coordinate space using the given camera parameters.  
To generate both training and test sequences, we translated a sliding window of length $T$ by one frame. Hence there is an overlap between the sequences. This gives us more data to train on, which is always an advantage for deep learning systems. During test time, we initially predict the first $T$ frames of the sequence and slide the window by a stride length of 1 to predict the next frame using the previous frames.
\paragraph{Training details} 
We trained our network for 100 epochs, where each epoch makes a complete pass over the entire Human 3.6M dataset. We used the Adam~\cite{adam} optimizer for training the network with a learning rate of $1e-5$ which is decayed exponentially per iteration. The weights of the LSTM units are initialized by Xavier uniform initializer~\cite{glorot2010understanding}. We used a mini-batch batch size of 32 i.e. 32 sequences. For most of our experiments we used a sequence length of 5, because it allows faster training with high accuracy. We experimented with different sequence lengths and found sequence length 4, 5 and 6 to generally give better results, which we will discuss in detail in the results section. We trained a single model for all the action classes. Our code is implemented in Tensorflow. We perform cross-validation on the training set to select the hyper-parameter values $\alpha$ and $\beta$ of our loss function to $1$ and $5$ respectively. Similarly, using cross-validation, the three hyper-parameters of the temporal consistency constraint $\eta,\rho$ and $\tau$, are set to $1, 2.5$ and $4$ respectively. A single training step for sequences of length 5 takes only 34 ms approximately, while a forward pass takes only about 16ms on NVIDIA Titan X GPU. Therefore given the 2D joint locations from a pose detector, our network takes about 3.2ms to predict 3D pose per frame.

\subsection{Quantitative results}
\paragraph{Evaluation on estimated 2D pose}

As mentioned before, we used a sequence length of 5 to perform both qualitative and quantitative evaluation of our network. The results on Human3.6M dataset~\cite{h36m_pami} under protocol \#1 are shown in Table~\ref{tab:protocol_1}. From the table we observe that our model achieves the lowest error for every action class under protocol \#1, unlike many of the previous state-of-the-art methods. Note that we train a single model for all the action classes unlike many other methods which trained a model for each action class.  Our network significantly improves the state-of-the-art result of Sun et al.~\cite{sun2017compositional} by approximately $12.1\%$ (by $7.2$ mm). The results under  protocol \#2, which aligns the predictions to the ground truth using a rigid body similarity transform before computing the error, is reported in Table~\ref{tab:protocol_2}. Our network improves the reported state-of-the-art results by $8.09\%$ (by $3.7$ mm) and achieves the lowest error for each action in protocol \#2  as well. From the results, we observe the effectiveness of exploiting temporal information across multiple sequences. By using the information of temporal context, our network reduced the overall error in estimating 3D joint locations, especially on actions like  \emph{phone},  \emph{photo}, \emph{sit} and \emph{sitting down} on which most previous methods did not perform well due to heavy occlusion. We also observe that our method outperforms both the baselines by a large margin on both the protocols. This shows that our method learned the temporal context of the sequences and predicted temporally consistent 3D poses, which naive post-processing techniques like temporal mean and median filters over frame-wise prediction failed to do. 

Like most previous methods, we report the results on action classes \emph{Walking} and \emph{Jogging} of the HumanEva~\cite{heva} dataset in Table~\ref{tab:heva}. We obtained the lowest error in four of the six cases and the lowest average error for the two actions. We also obtained the second best result on subject 2 of action \emph{Walking}. However, HumanEva is a smaller dataset than Human3.6M and the same subjects appear in both training and testing.

\begin{table}
\centering
\small
\hspace{-3mm}
\tabcolsep=1.0mm
\scalebox{0.65}{
\begin{tabular}{@{}l |lll |lll |l @{}}
\hline
& \multicolumn{3}{c}{Walking} & \multicolumn{3}{c}{Jogging} &\\
& S1 & S2 & S3 & S1 & S2 & S3 & Avg\\
\hline
Radwan et al.~\cite{radwan2013monocular}   & 75.1 & 99.8 & 93.8 & 79.2 & 89.8 & 99.4 & 89.5\\
Wang et al.~\cite{wang2014robust}          & 71.9 & 75.7 & 85.3 & 62.6 & 77.7 & 54.4 & 71.3\\
Simo-Serra et al.~\cite{simo2013joint}     & 65.1 & 48.6 & 73.5 & 74.2 & 46.6 & 32.2 & 56.7\\
Bo et al.~\cite{bo2010twin}                & 46.4 & 30.3 & 64.9 & 64.5 & 48.0 & 38.2 & 48.7\\
Kostrikov et al.~\cite{kostrikov2014depth} & 44.0 & 30.9 & 41.7 & 57.2 & 35.0 & 33.3 & 40.3\\
Yasin et al.~\cite{yasin2016dual}          & 35.8 & 32.4 & 41.6 & 46.6 & 41.4 & 35.4 & 38.9\\
Moreno-Noguer~\cite{distance-matrix}      & {19.7} & {\bf13.0} & {\bf24.9} & 39.7 & 20.0 & 21.0 & 26.9\\
Pavlakos et al.~\cite{volumetric}          & 22.1 & 21.9 & \underline{29.0} & 29.8 & 23.6 & 26.0 & 25.5\\
Lin et al~\cite{linCVPR17RPSM}			 & 26.5 & 20.7 & 38.0 & 41.0 & 29.7 & 29.1 & 30.8\\
Martinez et al.~\cite{JMartinez:ICCV:2017}  & {19.7} & 17.4 & 46.8 & \underline{26.9} & {18.2} & {18.6} & {24.6}\\
Fang et al.~\cite{fang2017learning}  & \underline{19.4} & 16.8 & 37.4 & {30.4} & \underline{17.6} & \underline{16.3} & \underline{22.9}\\
Ours  & \bf{19.1} & \underline{13.6} & 43.9 & {\bf23.2} & {\bf16.9} & {\bf15.5} & {\bf22.0}\\
\hline
\end{tabular}
}
\vspace{3mm}
\caption{Results on the HumanEva~\cite{heva} dataset, and comparison with previous work. The bold-faced numbers represent the best result while underlined numbers represent the second best.
}
\vspace{-10mm}
\label{tab:heva}
\end{table}

\paragraph{Evaluation on 2D ground truth}
As suggested by Martinez et al.~\cite{JMartinez:ICCV:2017}, we also found that the more accurate the 2D joint locations are, the better are the estimates for 3D pose. We trained our model on ground truth 2D poses for a sequence length of 5. The results under protocol \#1  are reported in Table~\ref{tab:protocol_1}.
As seen from the table, our model improves the lower bound error  of Martinez et al.~\cite{JMartinez:ICCV:2017} by almost $13.8\%$.

The results on ground truth 2D joint input for protocol \#2 are reported in Table~\ref{tab:roubst_gt}. When there is no noise in 2D joint locations, our network performs better than the models by Martinez et al.~\cite{JMartinez:ICCV:2017} and Moreno-Nouguer~\cite{distance-matrix}. These results suggest that the information of temporal consistency from previous frames is a valuable cue for the task of estimating 3D pose even when the detections are noise free.
\begin{figure*}[t]
\begin{center}
\includegraphics[width=\linewidth]{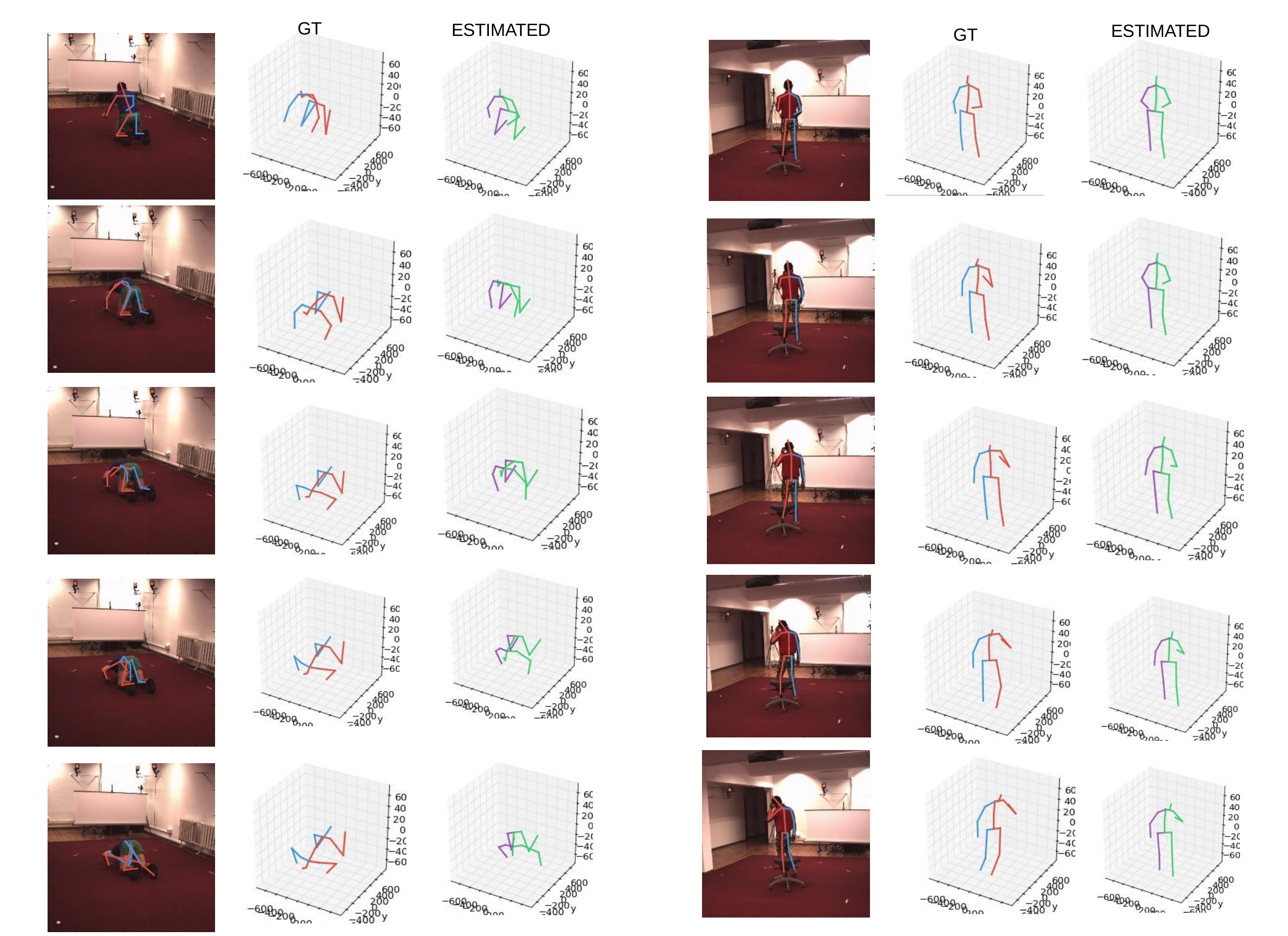}
\end{center}
\caption{Qualitative results on Human3.6M videos. The images on the \textbf{left} are for subject 11 and action \emph{sitting down}. On the \textbf{right} the images are for subject 9 and action \emph{phoning}. 3D poses in the center is the ground truth and on the right is the estimated 3D pose.}
\vspace{-5mm}
\label{fig:qualt2}  
\end{figure*}

\begin{figure*}
\begin{center}
\scalebox{1}{
\includegraphics[width=\linewidth,height=0.6\linewidth]{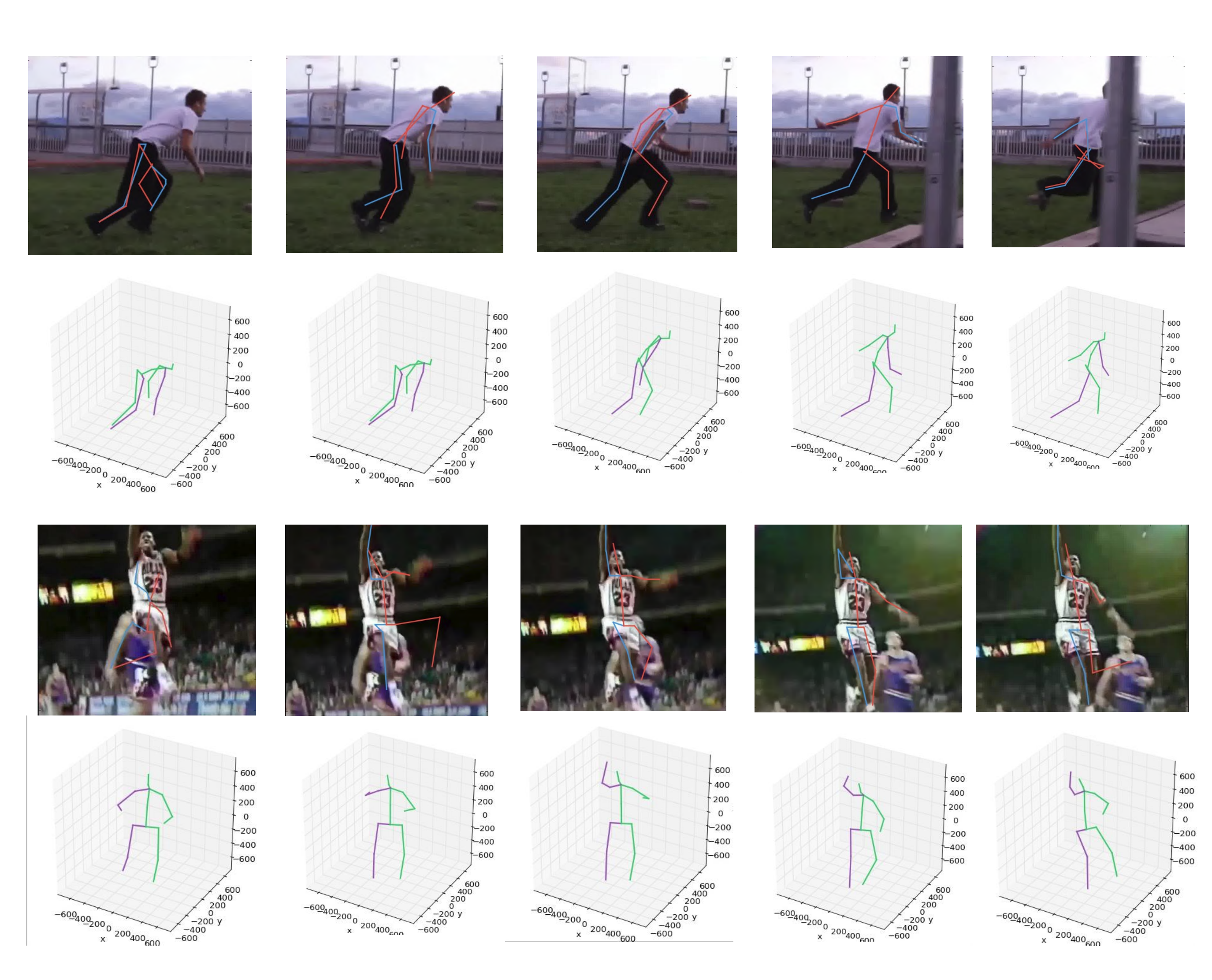}}
\caption{Qualitative results on Youtube videos. Note on the sequence at the top, our network managed to predict meaningful 3D poses even when the 2D pose detections were poor using temporal information of the past.}
\vspace{-5mm}
\end{center}
\label{fig:qual1} 
\end{figure*}

\begin{table}
\begin{center}
\tabcolsep=1mm
\scalebox{0.70}
{\begin{tabular}{@{}lccc@{}}
\hline
 & Moreno-Nouguer~\cite{distance-matrix} & Martinez et al.~\cite{JMartinez:ICCV:2017}& $Ours$\\
\hline
GT/GT                       & 62.17 & 37.10 & \textbf{31.67}\\
GT/GT + $\mathcal{N}(0,5)$  & 67.11  & 46.65 & \textbf{37.46}\\
GT/GT + $\mathcal{N}(0,10)$ & 79.12 & 52.84 & \textbf{49.41}\\
GT/GT + $\mathcal{N}(0,15)$ & 96.08 & \textbf{59.97} & 61.80\\
GT/GT + $\mathcal{N}(0,20)$ & 115.55 & \textbf{70.24} & 73.65\\
\hline
\end{tabular}}
\end{center}
\caption{Performance of our system trained with ground truth 2D pose of Human3.6M~\cite{h36m_pami} dataset and tested with different levels of additive Gaussian noise \textbf{(Top)} and on 2D pose predictions from stacked-hourglass~\cite{stacked-hourglass} pose detector \textbf{(Bottom)}under protocol \#2.} 
\vspace{-5mm}
\label{tab:roubst_gt}
\end{table}
\vspace{-7mm}
\paragraph{Robustness to noise}
We carried out some experiments to test the tolerance of our model to different levels of noise in the input data by training our network on 2D ground truth poses and testing on inputs corrupted by different levels of Gaussian noise. Table~\ref{tab:roubst_gt} shows how our final model compares against the models by  Moreno-Nouguer~\cite{distance-matrix} and Martinez et al.~\cite{JMartinez:ICCV:2017}. Our network is significantly more robust than Moreno-Nouguer's model~\cite{distance-matrix}. When compared against Martinez et al.~\cite{JMartinez:ICCV:2017} our network performs better when the level of input noise is low i.e. standard deviation less than or equal to 10. However, for higher levels of noise our network performs slightly worse than Martinez  et al.~\cite{JMartinez:ICCV:2017}. We would like to attribute the cause of this to the temporal smoothness constraint imposed during training which distributes the error of individual frames over the entire sequence. However, its usefulness can be observed in the qualitative results (See Figure~\ref{fig:qual1} and Figure~\ref{fig:qualt2}).
\begin{table}
\centering
\scalebox{0.65}{
\begin{tabular}{@{}lrr@{}}
\hline
 & error (mm) & $\Delta$\\
\hline
Ours & 51.9 & --\\
w/o weighted joints & 52.3 & 0.4\\
w/o temporal consistency constraint & 52.7 & 0.8\\
w/o recurrent dropout & 58.3 & 6.4\\
w/o layer normalized LSTM & 61.1 & 9.2\\
w/o layer norm and recurrent dropout & 59.5 & 7.6\\
w/o residual connections &  102.4 & 50.5\\
\hline
w non-fine tuned SH\cite{stacked-hourglass} & 55.6 & 3.7 \\
w CPM detections\cite{cpm} (14 joints) & 66.1 & 14.2 \\
\hline
\end{tabular}
}
\vspace{1.5mm}
\caption{Ablative and hyperparameter sensitivity analysis.}
\vspace{-5mm}
\label{tab:ablative_final}
\end{table}

\paragraph{Ablative analysis}
To show the usefulness of each component and design decision of our network, we perform an ablative analysis. We follow protocol \#1 for performing ablative analysis and trained a single model for all the actions. The results are reported in Table~\ref{tab:ablative_final}. We observe that the biggest improvement in result is due the the residual connections on the decoder side, which agrees with the hypothesis of He et al.~\cite{he2016deep}. Removing the residual connections massively increases the error by $50.5$ mm. When we do not apply layer normalization on LSTM units, the error increases by $9.2$ mm. On the other hand when dropout is not performed, the error raises by $6.4$ mm. When both layer normalization and recurrent dropout are not used the results get worse by $7.6$ mm. Although the temporal consistency constraint may seem to have less impact (only $0.8$ mm) quantitatively on the performance of our network, it ensures that the predictions over a sequence are smooth and temporally consistent which is apparent from our qualitative results as seen in Figure~\ref{fig:qual1} and Figure~\ref{fig:qualt2}. 

To show the effectiveness of our model on detections from different 2D pose detectors, we also experimented with the detections from CPM~\cite{cpm} and from stacked-hourglass~\cite{stacked-hourglass} (SH) module which is not fine-tuned on Human3.6M dataset. We observe that even for the non-fine tuned stacked hourglass detections, our model achieves the state-of-the-art results. For detections from CPM, our model achieves competitive accuracy for the predictions.

\begin{figure}[t]
\begin{center}
\includegraphics[width=0.50\linewidth, height=0.30\linewidth]{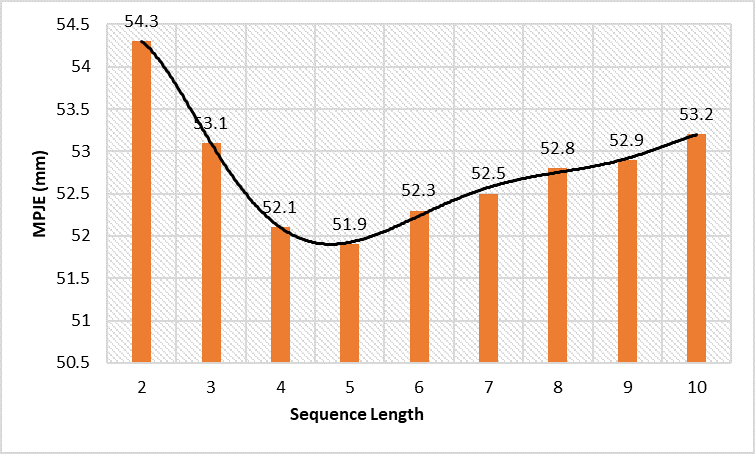}
\end{center}
\caption{Mean Per Joint Error(MPJE) in mm of our network for different sequence lengths.}
\label{fig:chart}  
\end{figure}

\paragraph{Performance on different sequence lengths}
The results reported so far have been for input and output sequences of length 5. We carried out experiments to see how our network performs for different sequence lengths ranging from 2 to 10. The results are shown in Figure~\ref{fig:chart}. As can be seen, the performance of our network remains stable for sequences of varying lengths. Even for a sequence length of 2, which only considers the previous and the current frame, our model generates very good results. Particularly the best results were obtained for length 4, 5 and 6. However, we chose sequence length 5 for carrying out our experiments as a compromise between training time and accuracy.


\subsection{Qualitative Analysis}
We provide qualitative results on some videos of Human3.6M and Youtube. We apply the model trained on the Human3.6M dataset on some videos gathered from Youtube, The bounding box for each person in the Youtube video is labeled manually and for Human3.6M the ground truth bounding box is used. The 2D poses are detected using the stacked-hourglass model fine-tuned on Human3.6M data. The qualitative result for Youtube videos is shown in Figure~\ref{fig:qual1} and for Human3.6M in Figure~\ref{fig:qualt2}. The real advantage of using the temporal smoothness constraint during training is apparent in these figures. For Figure~\ref{fig:qual1}, we can see that even when the 2D pose estimator breaks or generates extremely noisy detections, our system can recover temporally coherent 3D poses by exploiting the temporal consistency information. A similar trend can also be found for Human3.6M videos in Figure~\ref{fig:qualt2}, particularly for the action \emph{sitting down} of subject 11. We have provided more qualitative results in the supplementary material.

\section{Conclusion}
Both the quantitative and qualitative results for our network show the effectiveness of exploiting temporal information over multiple sequences to estimate 3D poses which are temporally smooth. Our network achieved the best accuracy till date on all of the 15 action classes in the Human3.6M dataset~\cite{h36m_pami}. Particularly, most of the previous methods struggled with actions which have a high degree of occlusion like  \emph{taking photo}, \emph{talking on the phone}, \emph{sitting} and \emph{sitting down}. Our network has significantly better results on these actions. Additionally we found that our network is reasonably robust to noisy 2D poses. Although the contribution of temporal smoothness constraint is not apparent in the ablative analysis in Table~\ref{tab:ablative_final}, its effectiveness is clearly visible in the qualitative results, particularly on challenging Youtube videos (see Figure~\ref{fig:qual1}). 

Our network effectively demonstrates the power of using temporal context information which we achieved using a sequence-to-sequence network that can be trained efficiently in a reasonably quick time. Also our network makes predictions from 2D poses at 3ms per frame on average which suggests that, given the 2D pose detector is real time, our network can be applied in real-time scenarios.

\clearpage

\bibliographystyle{splncs}
\bibliography{egbib}
\end{document}